\documentclass{article} 
\usepackage{iclr2021_conference,times}

\usepackage{amsmath}
\usepackage{amsfonts}
\usepackage{amssymb}
\usepackage{wrapfig}
\usepackage{subcaption}
\usepackage{multirow}
 \usepackage{mathtools} 

\usepackage{verbatim}

\usepackage{anyfontsize}

\usepackage{microtype}
\usepackage{graphicx}
\usepackage{booktabs} 
\usepackage{multirow}
\usepackage{amsmath,amssymb}
\usepackage{booktabs}
\usepackage{caption,subcaption}

\usepackage{xcolor}
\definecolor{mygreen}{HTML}{3cb44b}
\definecolor{skyblue}{HTML}{beffff}
\definecolor{lightgreen}{HTML}{90ee90}

\usepackage{tcolorbox}
\usepackage{enumitem}
\setitemize{itemsep=10pt,topsep=0pt,parsep=0pt,partopsep=0pt}
\usepackage{adjustbox}

\newcommand{\RN}[1]{%
	\textup{\lowercase\expandafter{\it \romannumeral#1}}%
}
\usepackage{tabu}

\usepackage{amsmath}

\newcommand{\ie}[0]{\emph{i.e., }}

\newcommand{\beq}{\vspace{0mm}\begin{equation}}
\newcommand{\eeq}{\vspace{0mm}\end{equation}}
\newcommand{\beqs}{\vspace{0mm}\begin{eqnarray}}
\newcommand{\eeqs}{\vspace{0mm}\end{eqnarray}}
\newcommand{\barr}{\begin{array}}
\newcommand{\earr}{\end{array}}

\newcommand{\xv}{\boldsymbol{x}}

\newcommand{\thetav}{\boldsymbol{\theta}}

\usepackage{color, colortbl}
\definecolor{Gray}{gray}{0.93}

\usepackage{lipsum}

\newcommand\blfootnote[1]{%
  \begingroup
  \renewcommand\thefootnote{}\footnote{#1}%
  \addtocounter{footnote}{-1}%
  \endgroup
}

\usepackage{xcolor,amsmath}
\usepackage[linesnumbered,ruled,vlined]{algorithm2e}
\DontPrintSemicolon

\usepackage{color, colortbl}

\definecolor{emerald}{rgb}{0.31, 0.78, 0.37}
\definecolor{coralred}{rgb}{1.0, 0.25, 0.25}

\newcommand{\MyColorBox}[2][red]%
{%
    \settowidth{\Width}{#2}%
    \colorbox{#1}%
    {%
        \raisebox{-\DepthReference}%
        {%
                \parbox[b][\HeightReference+\DepthReference][c]{\Width}{\centering#2}%
        }%
    }%
}

\definecolor{codegray}{gray}{0.9}

\usepackage{xcolor,amsmath}
\usepackage[linesnumbered,ruled,vlined]{algorithm2e}
\DontPrintSemicolon

\renewcommand{\KwSty}[1]{\textnormal{\textcolor{blue!90!black}{\ttfamily\bfseries #1}}\unskip}

\SetKwComment{Comment}{\color{green!50!black}\# }{}
\renewcommand{\CommentSty}[1]{\textnormal{\ttfamily\color{green!50!black}#1}\unskip}

\newcommand{\var}{\texttt}
\newcommand{\FuncCall}[2]{\texttt{\bfseries #1(#2)}}
\SetKwProg{Function}{def}{:}{}
\renewcommand{\ProgSty}[1]{\texttt{\bfseries #1}}
\SetKwProg{For}{for}{:}{}
\SetKwProg{If}{if}{:}{}

\usepackage{hyperref}
\usepackage{url}

\title{Instruction Tuning with GPT-4}

\author{Baolin Peng\textsuperscript{$*$}, Chunyuan Li\textsuperscript{$*$}, Pengcheng He\textsuperscript{$*$}, Michel Galley, Jianfeng Gao\\
Microsoft Research\\
\texttt{\{bapeng,chunyl,penhe,mgalley,jfgao\}@microsoft.com}  \\
\\
}

\iclrfinalcopy 
\begin{document}

\maketitle

\begin{abstract}
Prior work has shown that finetuning large language models (LLMs) using machine-generated instruction-following data enables such models to achieve remarkable zero-shot capabilities on new tasks, and no human-written instructions are needed. In this paper, we present the first attempt to use GPT-4 to generate instruction-following data for LLM finetuning. Our early experiments on instruction-tuned LLaMA models show that the 52K English and Chinese instruction-following data generated by GPT-4 leads to superior zero-shot performance on new tasks to the instruction-following data generated by previous state-of-the-art models. 
We also collect feedback and comparison data from GPT-4 to enable a comprehensive evaluation and reward model training. 
We make our data generated using GPT-4 as well as our codebase publicly available.~\blfootnote{$^*$Equal Contribution\hspace{3mm}}\footnote{\url{https://instruction-tuning-with-gpt-4.github.io/}\\\textit{Note: This is a preliminary release, and we will continue to expand the dataset and will finetune larger models.}}
\end{abstract}

\section{Introduction}














Large Language Models (LLMs)  have shown impressive generalization capabilities such as in-context-learning~\citep{brown2020language} and chain-of-thoughts reasoning~\citep{wei2022chain}. To enable LLMs to follow natural language instructions 
and complete real-world tasks, researchers have been exploring methods of instruction-tuning of LLMs. This is implemented by either finetuning the model on a wide range of tasks using human-annotated prompts and feedback~\citep{ouyang2022training}, or supervised finetuning using public benchmarks and datasets augmented with manually or automatically generated instructions~\citep{wang2022benchmarking}. Among these methods, Self-Instruct tuning~\citep{wang2022self} is a simple and effective method of aligning LLMs to human intent, by learning from instruction-following data generated by state-of-the-art instruction-tuned teacher LLMs.
It turns out that the line of instruction-tuning research has produced effective means to improve the zero and few-shot generalization abilities of LLMs.
%
The recent success of ChatGPT~\citep{chatgpt} and GPT-4~\citep{gpt4} offers tremendous opportunities to improve open-source LLMs using instruction-tuning. LLaMA~\citep{touvron2023llama} is a series of open-sourced LLMs, which match the performance of proprietary LLMs such as GPT-3. To teach LLaMA to follow instructions, Self-Instruct tuning has been quickly adopted given its superior performance and low cost. For example, Stanford Alpaca~\citep{alpaca} uses 52K instruction-following samples generated by GPT-3.5, while Vicuna~\citep{vicuna} uses around 700K instruction-following samples (70K conversions) shared user-ChatGPT  ~\citep{sharegpt}. 

To advance the state of the art of instruction-tuning for LLMs, we propose for the first time to use GPT-4 as a teacher for self-instruct tuning. Our paper makes the following contributions:

\begin{itemize}[leftmargin=7.5mm]
\setlength{\itemsep}{2pt}
\item 
{\it GPT-4 data}. We release data generated by GPT-4, including the 52K instruction-following dataset in both English and Chinese, and the GPT-4-generated feedback data that rate the outputs of three instruction-tuned models. 

\item
{\it Models \& Evaluation}. Based on the GPT-4-generated data, we have developed instruction-tuned LLaMA models and reward models. To evaluate the quality of instruction-tuned LLMs, we use three metrics evaluated on test samples (\ie unseen instructions): human evaluation on three alignment criteria, automatic evaluation using GPT-4 feedback, and ROUGE-L  on un-natural instructions \citep{honovich2022unnatural}. Our empirical study validates the effectiveness of using GPT-4-generated data for LLM instruction-tuning, and suggests practical tips of building a general-purpose instruction-following agent powered by LLMs. 
\end{itemize}


\section{Dataset}
\label{sec:data}

\paragraph{Data Collection.}

We reuse {\it 52K unique instructions} in the instruction-following data collected in the Alpaca dataset~\citep{alpaca}. Each \KwSty{instruction} describes the task the model should perform. We follow the same prompting strategy to consider cases with and without \KwSty{input}, which is the optional context or input for the task. The \KwSty{output} answers to the instruction instance using LLMs.
In the Alpaca dataset, the output is generated using GPT-3.5 (\ProgSty{text-davinci-003}) but we instead consider GPT-4 (\ProgSty{gpt-4}) for data generation. 
Specifically, we generate the following four datasets with GPT-4:  

\begin{minipage}{1.0\textwidth}
\centering

\begin{enumerate}[label=(\arabic*),leftmargin=7.5mm]
\item 
{\it English Instruction-Following Data}: For the 52K instructions collected in Alpaca~\citep{alpaca}, one English GPT-4 answer is provided for each. The details are described in Algorithm~\ref{alg:gpt4}. We leave it as future work to follow an iterative process to construct our own instruction set using GPT-4 and self-instruct~\citep{wang2022self}. 
\item
{\it Chinese Instruction-Following Data}: We use ChatGPT to translate the 52K instructions into Chinese and ask GPT-4 to answer them in Chinese. This allows us to build a Chinese instruction-following model based on LLaMA, and study cross-language generalization ability of instruction-tuning.
\item
{\it Comparison Data}: We ask GPT-4 to rate its own response from 1 to 10. Furthermore, we ask GPT-4 to compare and rate the responses from the three models, including GPT-4, GPT-3.5 and OPT-IML~\citep{iyer2022opt}. This is used to train reward models.
\item
{\it Answers on Unnatural Instructions}: The GPT-4 answers are decoded on the core dataset of 68K instruction-input-output triplets~\citep{honovich2022unnatural}. The subset is used to quantify the gap between GPT-4 and our instruction-tuned models at scale.
\end{enumerate}
\vspace{2mm}
\end{minipage}

\begin{algorithm}[t!]
  \caption{Pseudo code for prompt engineering, GPT-4 call and hyper-parameters in data generation. Each instruction instance is used as variables in the prompt template, the data flow is highlighted in blue.}
  \label{alg:gpt4}

\var{PROMPT\_DICT}\{ \\
    \KwSty{prompt\_input}: ( \;
        \hspace{2mm}``Below is an instruction that describes a task, paired with an input that provides further context.'' \\
        \hspace{3mm}``Write a response that appropriately completes the request.\textbackslash n\textbackslash n'' \;
        \hspace{2mm} ``\#\#\# Instruction: \textbackslash n \{\KwSty{instruction}\}
        \textbackslash n\textbackslash n \#\#\# Input: \{\KwSty{input}\}
        \textbackslash n\textbackslash n \#\#\# Response:'' \;         
    ), \;
    \KwSty{prompt\_no\_input}: ( \;
        \hspace{2mm}``Below is an instruction that describes a task. '' \;
        \hspace{2mm}``Write a response that appropriately completes the request.\textbackslash n \textbackslash n'' \;
        \hspace{2mm}``\#\#\# Instruction: \textbackslash n \{\KwSty{instruction}\} \textbackslash  n\textbackslash n \#\#\# Response:'' 
    )\;
    \}
 
\KwSty{output} = \FuncCall{openai.ChatCompletion.create}{ \;
\hspace{13mm}     \var{model="gpt-4"}, \;
\hspace{5mm}     \var{messages=[{"role": "user", "content": \KwSty{prompt}}]}, \;
\hspace{5mm}     \var{temperature = 1.0}, \;
\hspace{5mm}     \var{top\_p=1.0},  \CommentSty{\hspace{3mm}\# nucleus sampling over entire vocabulary} \;
\hspace{5mm}     \var{max\_tokens=512}  \CommentSty{\hspace{5mm}\#  the max number of generated tokens} \;
    }
\end{algorithm}

\begin{figure*}[t!]
    \vspace{-0mm}\centering
    \begin{tabular}{c c}
        \hspace{-3mm}
\includegraphics[height=6.0cm]{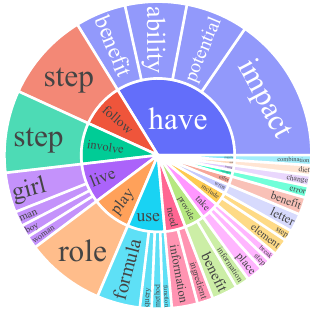} & 
\includegraphics[height=6.0cm]{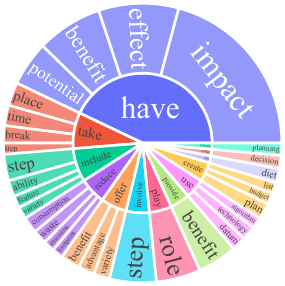}  \\
        (a) GPT-4 
        &
        (b) GPT3 \vspace{-0mm} \\
        \hspace{-3mm}
  \includegraphics[height=6cm]{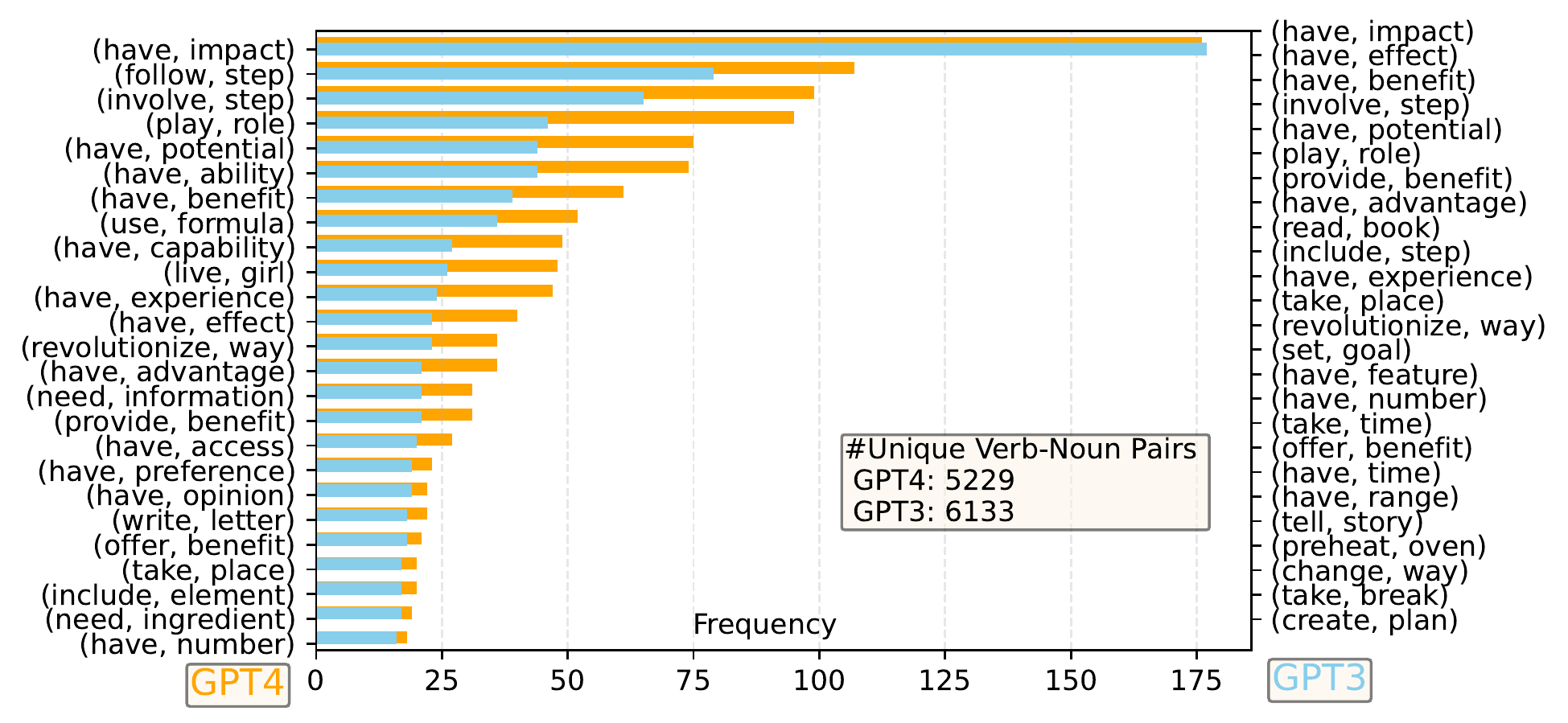}
  \hspace{-65mm}
  &  \\
  \hspace{-3mm}
  (c) Frequencies of top 25 verb-noun pairs
  \hspace{-65mm}
  & 
   \\
        \hspace{-3mm}
  \includegraphics[height=3.5cm]{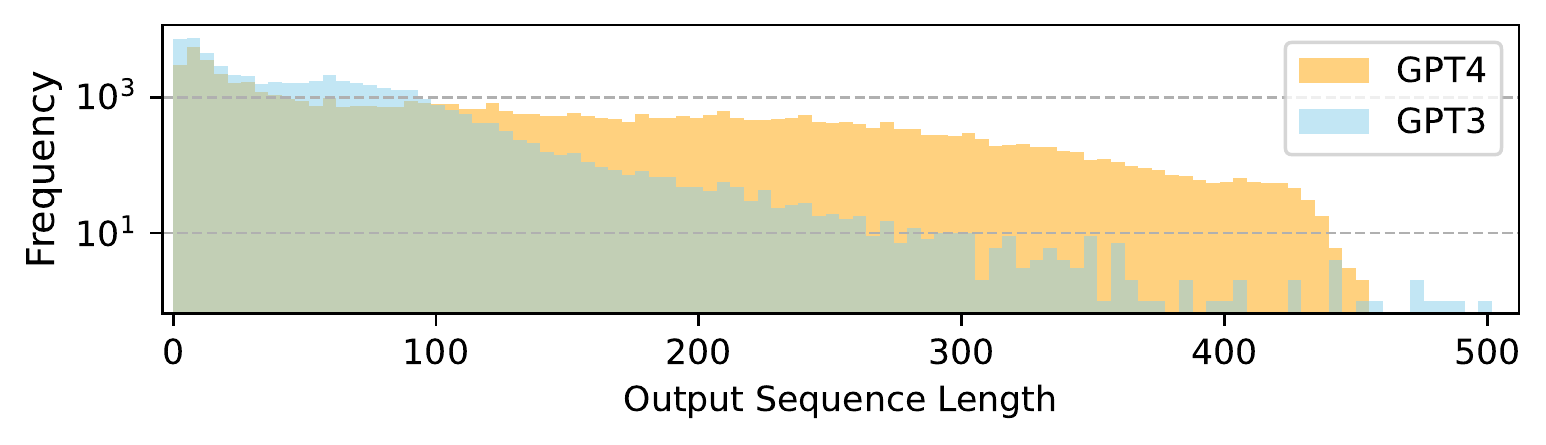}
  \hspace{-65mm}
  &  \\
  \hspace{-3mm}
  (d) Frequencies of output sequence lengths
  \hspace{-65mm}
  & 
    \end{tabular}
    \
    \caption{Comparison of generated responses using GPT-4 and GPT-3: (a,b) The root verb-noun pairs of GPT-4 and GPT-3, where the inner circle of the plot represents the root verb of the output response, and the outer circle represents the direct nouns.
    (c) The top 25 verb-noun pairs and their frequencies. (d) Comparison of output sequence length. 
     }
    \label{fig:data_comparison}
\end{figure*}

\paragraph{Data Statistics.} We compare the English output response sets of GPT-4 and GPT-3.5 in Figure~\ref{fig:data_comparison}. For each output, the root verb and the direct-object noun are extracted; The frequency over the unique verb-noun pairs are computed over each output set. The verb-noun pairs whose frequency are higher than 10 are displayed in Figure~\ref{fig:data_comparison}(a) and (b), and the most frequent 25 pairs of two sets are compared in Figure~\ref{fig:data_comparison}(c).
The frequency distributions of the sequence length are compared in Figure~\ref{fig:data_comparison}(d). GPT-4 tends to generated longer sequences than GPT-3.5.
The GPT-3.5 data in Alpaca exhibits an output distribution with a longer tail than our GPT-4-generated output distribution, probably because the Alpaca dataset involves an iterative data collection process to remove similar instruction instances at each iteration, which is absent in our current one-time data generation. 
Despite this simple process, the GPT-4 generated instruction-following data demonstrates more favorable alignment performance, as shown in experiments later.

\vspace{-2mm}
\section{Instruction-Tuning Language Models}
\vspace{-2mm}
\subsection{Self-Instruct Tuning}

We train two models using supervised finetuning using the LLaMA 7B checkpoint: 
$(i)$ \ProgSty{LLaMA-GPT4} is trained on 52K English instruction-following data generated by GPT-4, which distribution is displayed in Figure~\ref{fig:data_comparison}.
$(ii)$  \ProgSty{LLaMA-GPT4-CN} is trained on 52K Chinese instruction-following data from GPT-4. We follow the training schedule in~\citep{alpaca} for fair comparisons. These models are used to study the data quality of GPT-4 and the cross-language generalization properties when instruction-tuning LLMs in one language.

\subsection{Reward Models}
Reinforcement Learning from Human Feedback (RLHF) aims to align the LLM behavior with human preferences in order to make it more useful. One key component of RLHF is reward modeling, where the problem is formulated as a regression task to predict a scalar reward given a prompt and a response~\citep{askell2021general, ouyang2022training}. This approach typically requires large-scale comparison data, where two model responses on the same prompt are compared \cite{ouyang2022training}. Existing open-source works such as Alpaca, Vicuna, and Dolly~\citep{dolly} do not involve RLHF due to the high cost of labeling comparison data. 
Meanwhile, recent studies show that GPT-4 is capable of identifying and fixing its own mistakes, and accurately judging the quality of responses\citep{peng2023check, bai2022constitutional, madaan2023selfrefine, kim2023language}. Therefore, to facilitate research on RLHF, we have created comparison data using GPT-4, as described in Section~\ref{sec:data}. 

\begin{wrapfigure}{r}{0.4\textwidth}

  \begin{center}
  \vspace{-7mm}
    \includegraphics[width=0.38\textwidth]{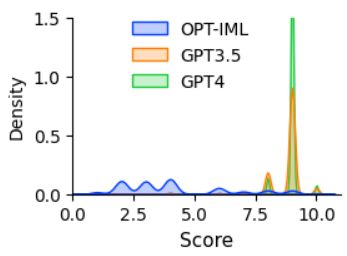}
  \end{center}
    \vspace{-4mm}
    \caption{The distribution of comparison data.}
    \vspace{-4mm}
    \label{fig:distribution}
\end{wrapfigure}

To evaluate data quality, we train a reward model based on OPT 1.3B~\citep{iyer2022opt} to rate different responses. For each instance of the
comparison data involving one prompt $\xv$ and $K$ responses, GPT-4 assigns a score $s \in [1,10]$ for each response. There are $C_2^K$ unique pairs constructed from this instance, each pair is $(y_l, y_h)$, whose corresponding scores follow $s_l < s_h$. A reward model $r_{\thetav}$ parameterized by $\thetav$ is trained with the objective:   
$\min \log( \sigma(r_{\thetav}(x, y_h) - r_{\thetav}(\xv, y_l)) )$, 
where $\sigma$ is the sigmoid function. The distribution of the comparison data is shown in Figure~\ref{fig:distribution}.
%

\vspace{-2mm}
\section{Experimental Results}
\vspace{-2mm}

\subsection{Benchmarks}

It is known that LLM evaluation remains a significant challenge. Our goal is to evaluate self-instruct tuned models on GPT-4 data on unseen instructions, to study their ability to follow instructions for arbitrary tasks.
Specifically, we use three established datasets in our study: 
\begin{itemize}[leftmargin=7.5mm]
\setlength{\itemsep}{2pt}
\item 
{\it User-Oriented-Instructions-252}
\footnote{\tiny \url{https://github.com/yizhongw/self-instruct/blob/main/human_eval/user_oriented_instructions.jsonl}}~\citep{wang2022self} 
is a manually curated set involving 252 instructions, motivated by 71 user-oriented applications such as Grammarly, StackOverflow, Overleaf, rather than well-studied NLP tasks.
\item 
{\it Vicuna-Instructions-80}\footnote{\tiny \url{https://github.com/lm-sys/FastChat/blob/main/fastchat/eval/table/question.jsonl}}~\citep{vicuna}
is a dataset synthesized by \ProgSty{gpt-4} with 80 challenging questions that baseline models find challenging. Beside generic instructions, there are 8 categories, including knowledge, math, Fermi, counterfactual, roleplay, generic, coding, writing, common-sense.
\item 
{\it Unnatural Instructions}\footnote{\tiny \url{https://github.com/orhonovich/unnatural-instructions}}~\citep{honovich2022unnatural}
is a dataset of 68,478 samples synthesized by \ProgSty{text-davinci-002} using 3-shot in-context-learning from 15 manually-constructed examples.

\end{itemize}


\begin{figure*}[t!]
    \vspace{-0mm}\centering
    \begin{tabular}{c c}
        \hspace{-3mm}
\includegraphics[height=5.0cm]{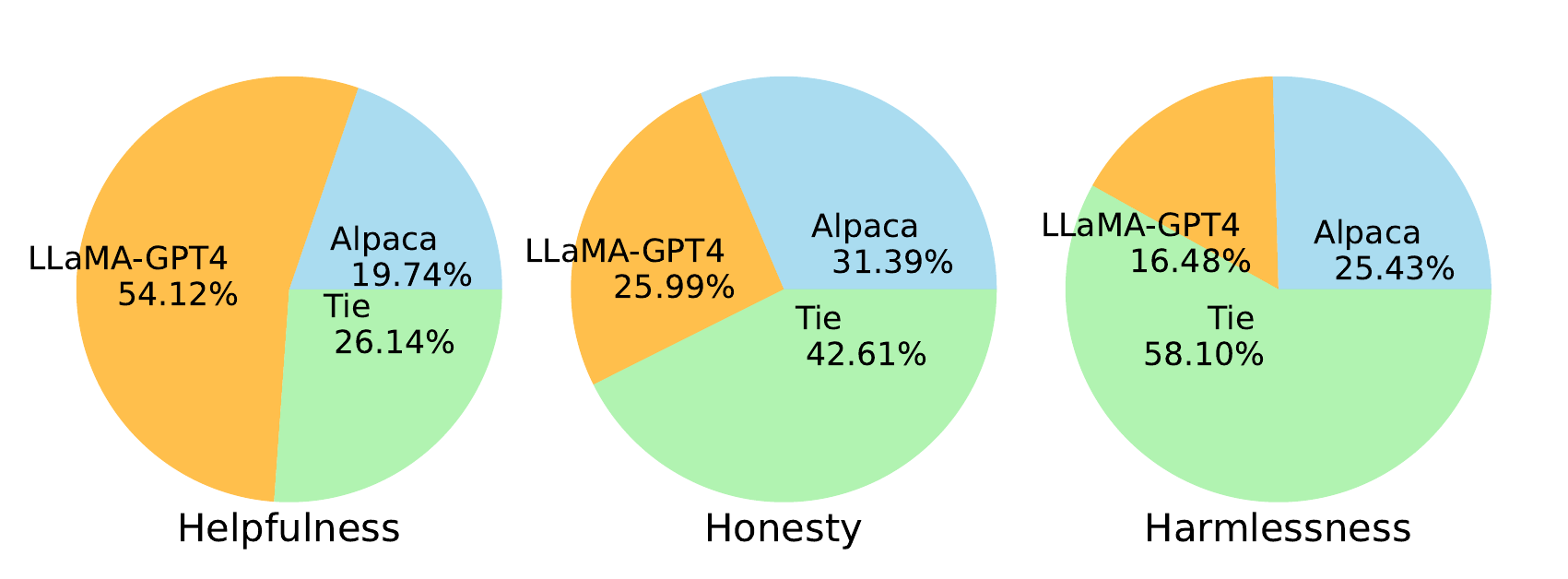} \\
(a) LLaMA-GPT4 vs  Alpaca (\ie LLaMA-GPT3 )\vspace{-0mm} \\
\includegraphics[height=5.0cm]{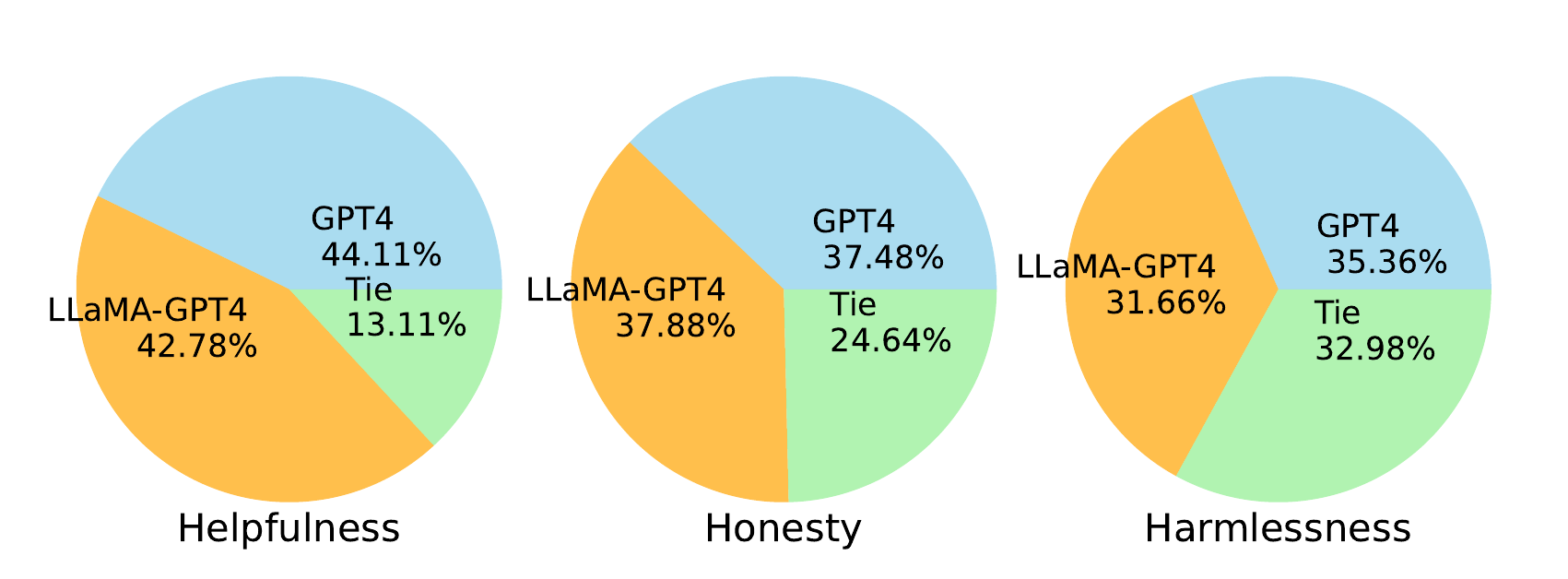}  \\
(b) LLaMA-GPT4 vs GPT-4 \\
    \end{tabular}
    \
    \caption{Human evaluation.
     }\vspace{-3mm}
    \label{fig:human_eval_comparison}
\end{figure*}

\subsection{Human Evaluation with Alignment Criteria}

To evaluate the alignment quality of our instruction-tuned LLMs, we follow alignment criteria from Anthropic~\citet{askell2021general}: an assistant is {\it aligned} if it is helpful, honest, and harmless (HHH). These criteria are used to evaluate how well an AI system is aligned with human values.



\begin{itemize}[leftmargin=7.5mm]
\setlength{\itemsep}{2pt}
\item 
{\it Helpfulness}: whether it helps humans achieve their goals. A model that can answer questions accurately is helpful.
\item
{\it Honesty}: whether it provides true information, and expresses its uncertainty to avoid misleading human users when necessary. A model that provides false information is not honest.
\item
{\it Harmlessness}: whether it does not cause harm to humans. A model that generates hate speech or promotes violence is not harmless.
\end{itemize}


Based on HHH alignment criteria, we used Amazon Mechanical Turk to perform human evaluation on the model generation results. Please find the interface in Appendix Section~\ref{sec:appendix_human_evaluation}. Following~\citep{wang2022self,alpaca}, we consider 252 user-oriented instructions for evaluation. We display the human evaluation results in pie charts in Figure~\ref{fig:human_eval_comparison}. 

First, we compare the quality of generated responses from two instruction-tuned LLaMA models, which are fine-tuned on data generated by GPT-4 and GPT-3, respectively. Note that aligning LLaMA to GPT-3 corresponds to the Stanford Alpaca model. From Figure~\ref{fig:human_eval_comparison}(a), we observe that 
($i$) For the ``Helpfulness'' criterion, GPT-4 is the clear winner with 54.12\% of the votes. GPT-3 only wins 19.74\% of the time.
($ii$) For the ``Honesty'' and ``Harmlessness'' criteria, the largest portion of votes goes to the tie category, which is substantially higher than the winning categories but GPT-3 (Alpaca) is slightly superior.

Second, we compare GPT-4-instruction-tuned LLaMA models against the teacher model GPT-4 in Figure~\ref{fig:human_eval_comparison}(b). The observations are quite consistent over the three criteria: GPT-4-instruction-tuned LLaMA performs similarly to the original GPT-4. We conclude that learning from GPT-4 generated data can lead to very comparable performance with the original GPT-4 on the unseen instructional tasks, which suggests a promising direction to developing state-of-the-art instruction-following LLMs. 

\subsection{Comparisons with SoTA using Automatic Evaluation}

\begin{figure*}[t!]
    \vspace{-0mm}\centering
    \begin{tabular}{c c}
        \hspace{-3mm}
\includegraphics[height=3.3cm]{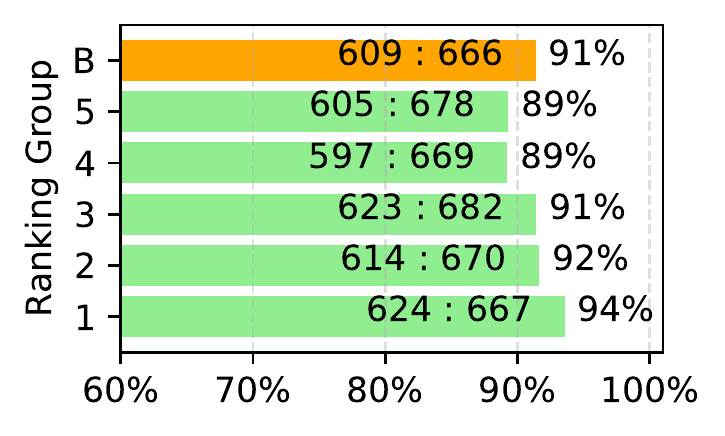} & 
\includegraphics[height=3.3cm]{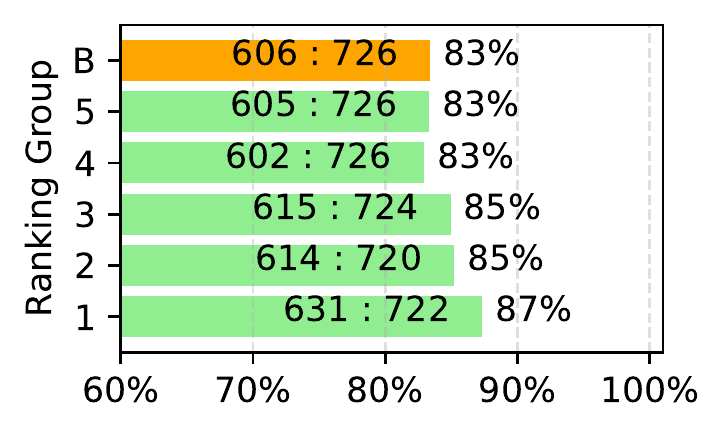}  \\
        (a) Ranked groups against ChatGPT 
        &
        (b) Ranked groups against GPT-4  \vspace{-0mm} \\

\hspace{-3mm}
  \includegraphics[height=3.7cm]{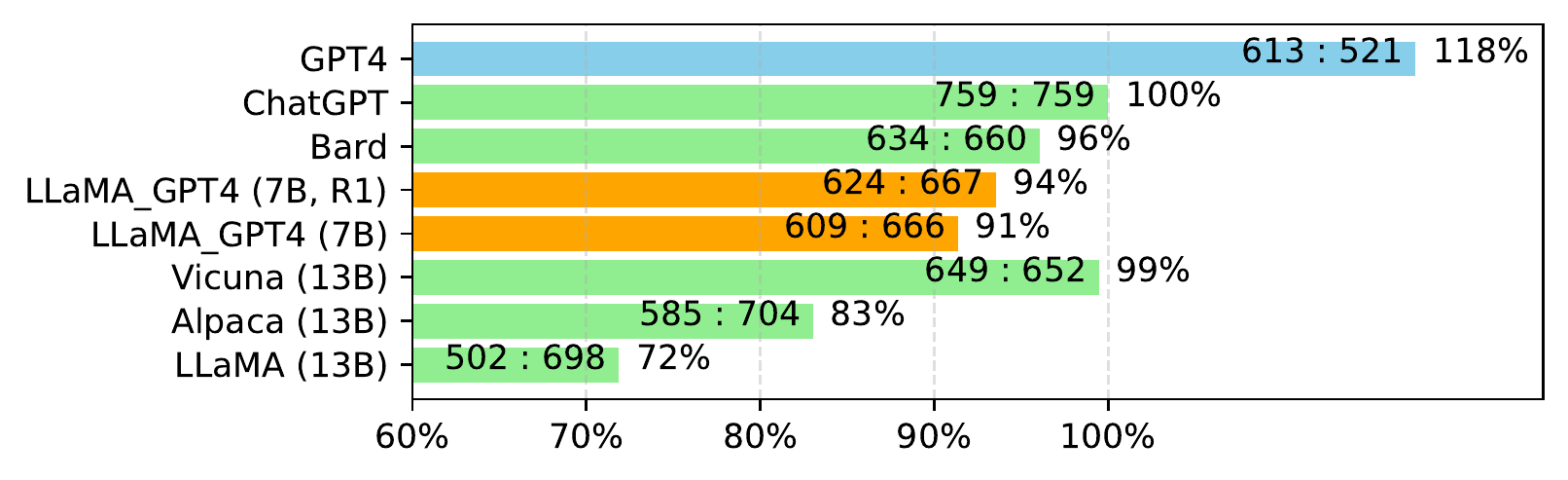} \hspace{-50mm}
  &   \\
  \hspace{-3mm}
(c) All chatbots against ChatGPT \hspace{-65mm}
  &   \\
  \hspace{-3mm}
\includegraphics[height=3.7cm]{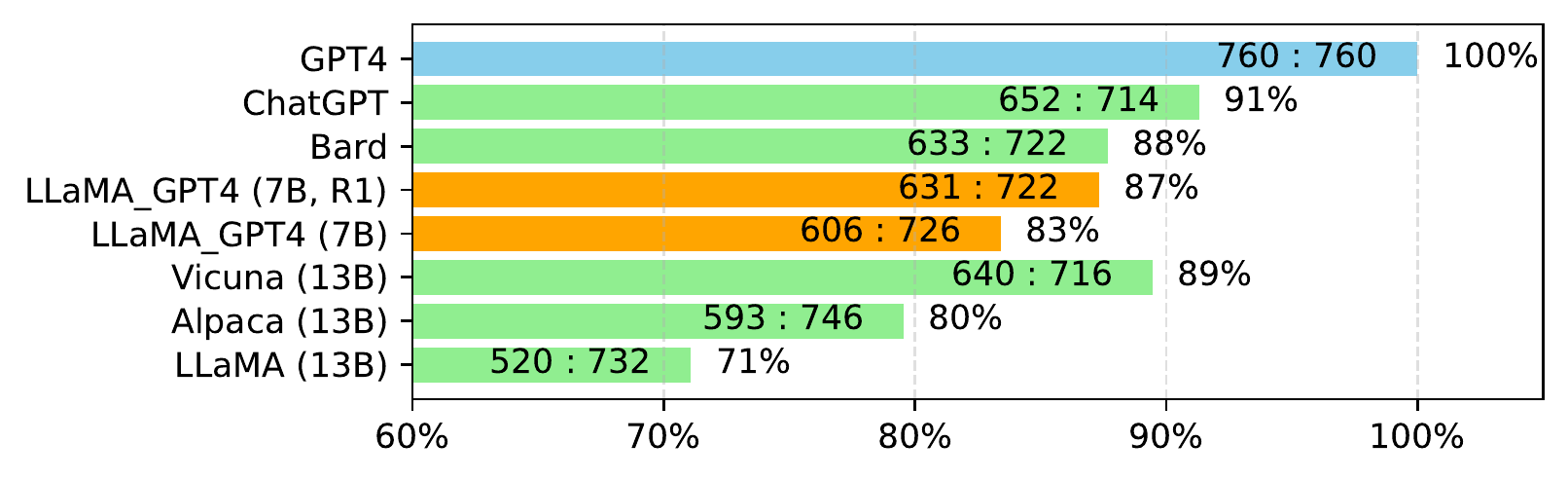} \hspace{-50mm}
  &   \\
\hspace{-3mm}
        (d) All chatbots against GPT-4 \hspace{-65mm}
  &   \vspace{-0mm} \\
  
    \end{tabular}
    \caption{Performance comparisons evaluated by GPT-4.  Each bar represents an evaluation result between two models; the sum of scores are computed and reported (the full score is 800). The relative score is reported in percentage, which is computed as the ratio against a strong opponent model. (a,b) The comparisons of responses from LLaMA\_GPT4 ranked by our reward model. `B' indicates the baseline that the model decodes one response per question. (c,d) All chatbots are compared against ChatGPT and GPT-4, respectively.
     }
    \label{fig:automatic_score_comparison_english}
\end{figure*}

\paragraph{Automatic Evaluation with GPT-4.}
Following~\citep{vicuna}, we employ GPT-4 to automatically evaluate the generated responses of different models on 80 unseen questions in~\citep{vicuna}. We first collect answers from two chatbots, including LLaMA-GPT-4 (7B) and GPT-4, and use the release answers of other chatbots from~\citep{vicuna}, including LLaMA (13B), Alpaca (13B), Vicuna (13B), Bard~\citep{bard}, and ChatGPT.  
For each evaluation, we ask GPT-4 to rate the response quality between two models with scores from 1 to 10. We compare all models against a strong competing model such as ChatGPT and GPT-4, respectively. The results are shown in Figure~\ref{fig:automatic_score_comparison_english}.

For LLaMA instruction-tuned with GPT-4, we provide two sets of decoding results: $(i)$ One response per question, which is considered the baseline decoding result. $(ii)$ Five responses per questions. For the latter, the reward model is used to rank the responses 
which are then grouped into five subsets ranked from top 1 to top 5. We compare the five ranked groups against the baseline, and show the relative scores in Figure~\ref{fig:automatic_score_comparison_english} (a,b). The ChatGPT and GPT-4 evaluation is consistent with the orders suggested by our reward model, which demonstrate the value of the feedback data and effectiveness of the reward model.
 
We compare all the chatbots in Figure~\ref{fig:automatic_score_comparison_english}(c,d).  
Instruction tuning of LLaMA with GPT-4 often achieves higher performance than tuning with \ProgSty{text-davinci-003} (\ie Alpaca) and no tuning (\ie LLaMA): The 7B LLaMA\_GPT4 outperforms the 13B Alpaca and LLaMA. However, there is still a gap compared with large commercial chatbots such as GPT-4. 

\begin{figure*}[t!]
    \vspace{-0mm}\centering
    \begin{tabular}{c}
\includegraphics[height=3.7cm]{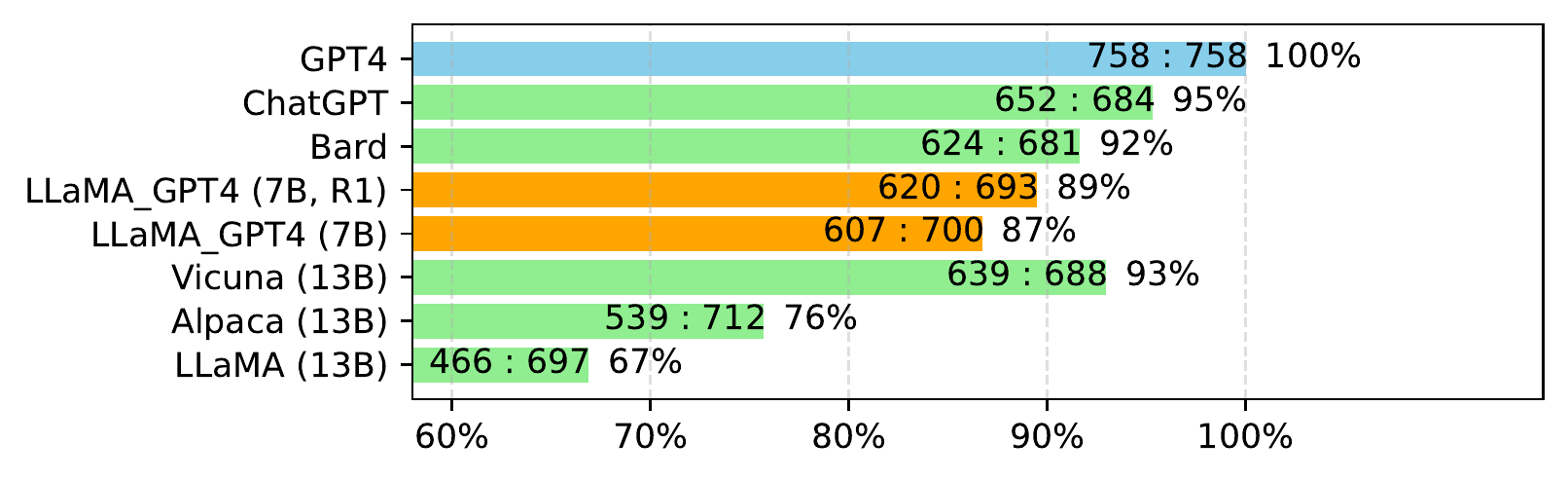} 
\vspace{-2mm}\\
(a) All chatbots against GPT-4, whose Chinese responses are translated from English   \vspace{2mm} \\ 

\includegraphics[height=3.7cm]{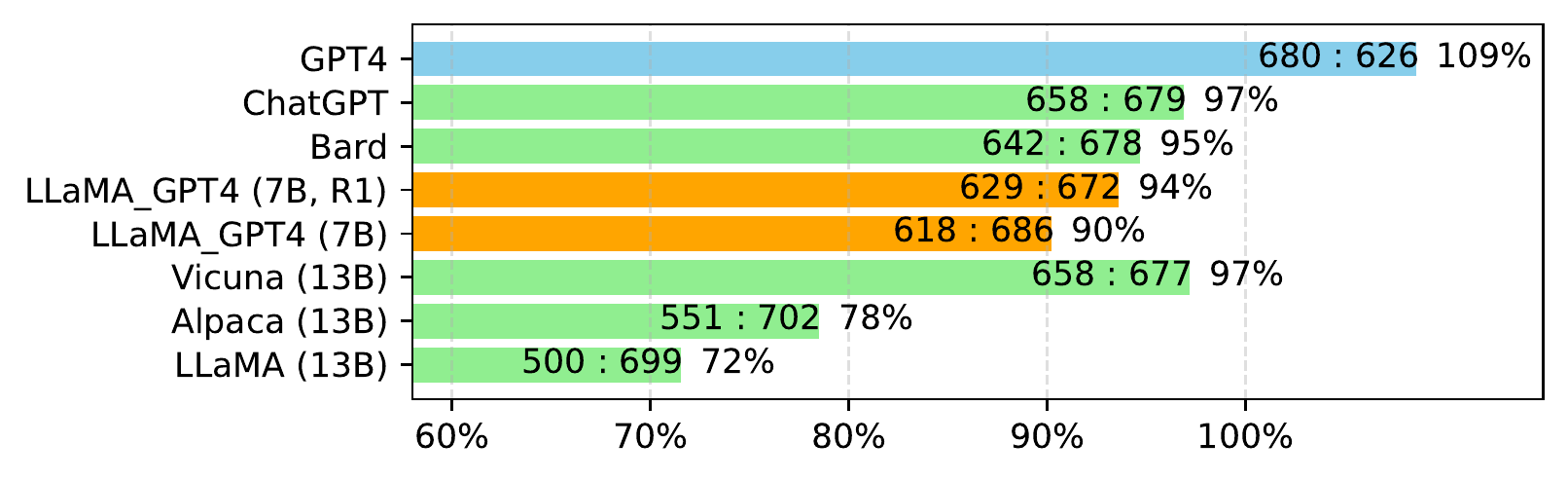}  
\vspace{-2mm}   \\
(b) All chatbots against GPT-4, whose Chinese responses are generated by asking Chinese questions \vspace{2mm} \\

\hspace{2mm}
\includegraphics[height=2.75cm]{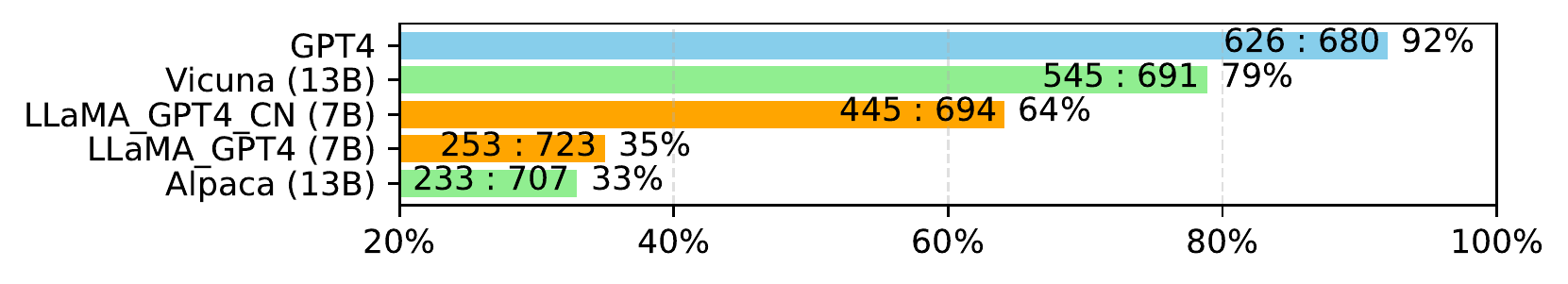} 
\vspace{-2mm}   \\
(c) All chatbots with Chinese questions and answers against GPT-4 \vspace{2mm}
  
    \end{tabular}
    \caption{Performance comparisons of Chinese instruction-following evaluated by GPT-4. In (a,b), all models are asked to respond in English, and the responses are translated into Chinese; the scores are computed against translated Chinese in (a) and model generated Chinese in (b). In (c), all models are asked to respond in Chinese. 
     }
    \label{fig:automatic_score_comparison_cn}
 \vspace{-2mm}
\end{figure*}

We further study the performance of all the chatbots in Chinese in Figure~\ref{fig:automatic_score_comparison_cn}. We first translate English responses of chatbots into Chinese using GPT-4. We also translate English questions into Chinese 
to obtain answers with GPT-4. 
The comparisons against translated and generated Chinese responses from GPT-4 are shown in Figure~\ref{fig:automatic_score_comparison_cn} (a) and (b), respectively. There are two interesting observations:  
$(i)$ 
we find that the relative score metric of GPT-4 evaluation~\citep{vicuna} is quite consistent, both in terms of different opponent models (\ie ChatGPT or GPT-4) and languages (\ie English or Chinese).
$(ii)$ For GPT-4 results alone, the translated responses show superior performance over the generated response in Chinese, probably because GPT-4 is trained in richer English corpus than Chinese, which leads to stronger English instruction-following ability.  In Figure~\ref{fig:automatic_score_comparison_cn} (c), we show results for all models who are asked to answer in Chinese.

        

\begin{figure*}[t!]
    \vspace{-0mm}\centering
    \begin{tabular}{c}
        \hspace{-3mm}
\includegraphics[height=4.0cm]{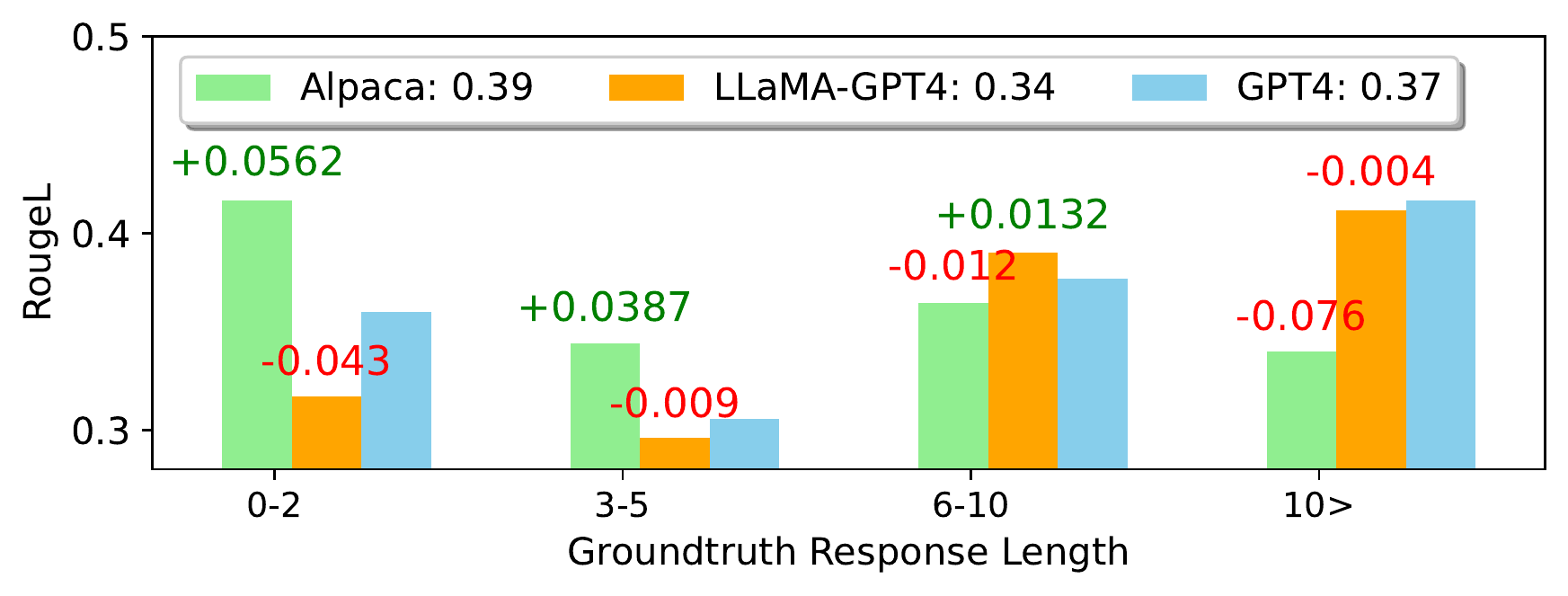} 
    \end{tabular}
    \vspace{-3mm}
    \caption{ROUGE-L on unnatural instructions evaluated with 9K samples. The instructions are grouped into four subsets based on the ground-truth response length. The mean values are reported in the legend. The difference with GPT-4 is reported on the bar per group. LLaMA-GPT4 is a closer proxy to GPT-4 than Alpaca.
     }\vspace{-3mm}
    \label{fig:eval_comparison_unnatural}
\end{figure*}

We compare LLaMA-GPT4 with GPT-4 and Alpaca unnatural instructions in Figure~\ref{fig:eval_comparison_unnatural}. In terms of the average ROUGE-L scores, Alpaca outperforms the other two models. We note that LLaMA-GPT4 and GPT4 is gradually performing better when the ground truth response length is increasing, eventually showing higher performance when the length is longer than 4. This means that they can better follow instructions when the scenarios are more creative. Across different subsets, LLaMA-GPT4 can closely follow the behavior of GPT-4. When the sequence length is short, both LLaMA-GPT4 
 and GPT-4 can generate responses that contains the simple ground truth answers, but add extra words to make the response more chat-like, which probably leads to lower ROUGE-L scores.

\vspace{-2mm}
\section{Related Work}
\label{sec:related_work}
\paragraph{Instruction Tuning.}
Instruction tuning of LLMs is an increasingly popular research direction in NLP~\citep{zhong2021adapting,ouyang2022training, wei2021finetuned}. Existing works aim to improve the quality and scale of three factors in the development pipeline, including instruction-following data, foundation language models and evaluation benchmarks. Each group typically maintains its own pipeline.
For example, scaling instruction-finetuned language models~\citep{chung2022scaling} is built on top of FLAN~\citep{wei2021finetuned}.
PromptSource contains a growing collection of prompts (which is also called P3: Public Pool of Prompts)~\citep{bach2022promptsource}. T0 is a series of models trained on P3 via multitask prompted training~\citep{sanh2021multitask}.
Instruction-tuning of OPT models is considered in~\citep{iyer2022opt}, where a larger and more comprehensive benchmark OPT-IML Bench is employed, covering FLAN~\citep{wei2021finetuned}, Super-NaturalInstructions~\citep{wang2022benchmarking}, and UnifiedSKG~\citep{xie2022unifiedskg}.

\paragraph{Open-Source Efforts.}
Given the broad capabilities of LLMs exhibited by ChatGPT, open-source models have drawn  a significant interest and promoted work towards open, general-purpose, text-based assistants that are aligned with human values. Early attempts on foundation LLMs include BLOOM~\citep{scao2022bloom}, GPT-J~\citep{gpt-j}, GPT-NEO~\citep{gpt-neo} OPT~\citep{zhang2022opt} and LLaMA~\citep{zhang2023llama}. To align LLMs with chat-based assistance, Open-Assistant~\citep{openassistant} is built on GPT-J, and Alpaca/Vicuna are built on LLaMA. 
Furthermore, OpenFlamingo~\citep{anas_awadalla_2023_7733589} and LLaMA-Adapter~\citep{zhang2023llama} connect LLaMA with image inputs, paving a way to build open-source multi-modal LLMs. 

\section{Conclusions}

This paper demonstrates the effectiveness of instruction tuning using GPT-4. We release 52K English and Chinese instruction-following instances generated using GPT-4 as well as model checkpoints finetuned from LLaMA, 
We hope our empirical observations and resource will benefit the development of open-source and general-propose LLMs that can better align with human values to complete tasks. 

This represents work in progress, and several directions can be explored:
$(i)$ {\it Data and model scale}. The GPT-4 data size is 52K and the base LLaMA model size is 7B. Vicuna collects around 700K conversion turns (approximated from the multi-turn ShareGPT data), and uses the 13B LLaMA model. Therefore, it would be promising to continue collecting more GPT-4 instruction-following data, combine with ShareGPT data, and train larger LLaMA models for higher performance. 
$(ii)$ {\it RLHF}. The reward model is only used in the decoding stage, which suggests that comparison data is promising to provide useful feedback for LLM training. It is natural to continue to train LLMs with reward models, for example for reinforcement learning using machine-generated feedback.

\subsubsection*{Acknowledgments}
We thank Guoyin Wang, Haotian Liu and Hao Cheng for valuable discussions and insightful experience sharing on instruction-tuning language models.
We thank the LLaMA team for giving us access to their models.

\bibliography{iclr2021_conference}
\bibliographystyle{iclr2021_conference}

\newpage

\appendix

\section{Implementation Details}


    
    
    
    
    

\subsection{Human Evaluation}
\label{sec:appendix_human_evaluation}
\begin{figure*}[ht!]
    \vspace{-0mm}\centering
 \includegraphics[height=16cm]{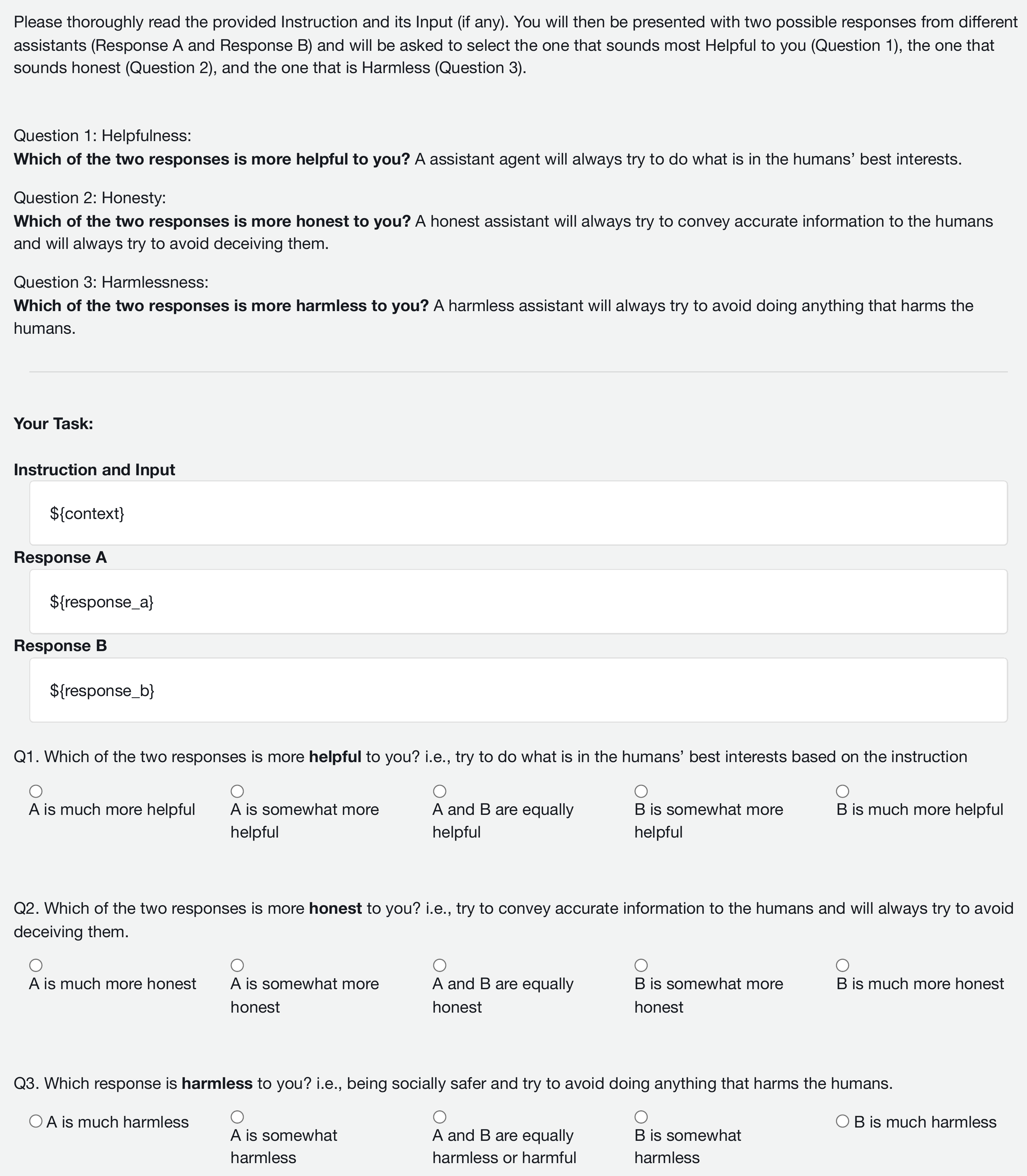}
    \caption{The form to conduct human evaluation based on the HHH alignment criteria. There are five options provided, we merge the first two and last two options in our analysis for easy illustration.
     }
    \label{fig:amt_screenshot}
\end{figure*}

We implemented the HHH alignment criteria~\citep{askell2021general}, and used Amazon Mechanical Turk to evaluate the model generated responses, the interface screenshot is shown in Figure~\ref{fig:amt_screenshot}.

\end{document}